\documentclass[11pt,a4paper]{article}
\usepackage[hyperref]{acl2018}
\usepackage{times}
\usepackage{latexsym}
\usepackage{balance}
\usepackage{graphicx}
\usepackage{url}

\aclfinalcopy % Uncomment this line for the final submission

\title{Hierarchical Text Generation using an Outline}

\author{Mehdi Drissi \\
  Harvey Mudd College \\ Claremont, CA, USA \\
  {\tt mdrissi@hmc.edu} \\\And
  Olivia Watkins \\
  Harvey Mudd College \\ Claremont, CA, USA \\
  {\tt owatkins@hmc.edu} \\\And 
  Jugal Kalita \\
  University of Colorado Springs \\ Colorado Springs, CO, USA \\
  {\tt jkalita@uccs.edu}
  }

\begin{document}
\maketitle
\begin{abstract}
Many challenges in natural language processing require generating text, including  language translation, dialogue generation, and speech recognition. For all of these problems, text generation becomes more difficult as the text becomes longer. Current language models often struggle to keep track of coherence for long pieces of text. Here, we attempt to have the model construct and use an outline of the text it generates to keep it focused. We find that the usage of an outline improves perplexity. We do not find that using the outline improves human evaluation over a simpler baseline, revealing a discrepancy in perplexity and human perception. Similarly, hierarchical generation is not found to improve human evaluation scores.
\end{abstract}

\section{Introduction}
Recurrent neural networks have been successfully used for a variety of tasks in dealing with natural language. Successes include language translation \cite{Sutskever2014}, speech recognition \cite{Graves2012}, and text to speech \cite{Simonyan}. They all learn to model the conditional probability of a sequence and generate words sequentially, typically using  sequence to sequence models. These models have the advantage that their negative log likelihood is differentiable, allowing them to be directly trained through gradient descent.

A similar task is language modeling. Here the goal is to determine the probability of a sequence of words. Being able to model text is important for natural language understanding. These models can be used for re-ranking candidate translations \cite{Cho2014}, determining possible ways to extend a piece of writing \cite{Roemmele2018}, and generating text \cite{Dauphin2016}. These problems all share the difficulty that although in theory a recurrent model can preserve information for arbitrarily long sequences, in practice recurrent models tend to struggle to keep track of context as the sequence length becomes large. Models for tasks like language translation tend to avoid translating entire paragraphs and instead focus on only generating a sentence at a time. Similarly, when one generates multiple sentences of text from language models, they tend to be locally coherent but not globally coherent.

The difficulty of generating large samples is not unique to text and also occurs with images. In the realm of images though, a different technique is commonly used for generation. Here, generative adversarial networks \cite{Goodfellow2014a} have been successfully used to generate images directly. Similar to text, initially these models only worked well for generating small images. Generating large images (like 1024 by 1024) was difficult for these models. In some recent work, the issue of large images was dealt with by generating images in a hierarchical manner. Instead of directly learning to generate the desired image, lower resolution images were first generated and then improved upon \cite{Zhang2016a}. Initially, this was done by generating one lower resolution image and then directly the final image. More recently, this has been extended to starting off with generating a small image and iteratively increasing its resolution by double until reaching the desired size \cite{Karras2017}. 

Inspired by the idea of generating images in a hierarchical manner, here we will explore generating text in a hierarchical manner. Similar to \cite{Zhang2016a}, we will approach this by generating in two steps. One difficulty with hierarchical text generation is that meaningfully downsampling text is more difficult than downsampling images. The textual equivalent of decreasing resolution is summarization, which is a difficult problem in itself. To side-step this issue, we will use a simple extractive summarization approach to get an outline. Using an extractive summarization approach to acquire information to build upon has been done previously in generating Wikipedia articles \cite{Liu2018}. Their approach differs from ours in that there they used summarization to extract relevant information from references to generate the article, while here we are summarizing the target text to acquire an outline of it. We generate complete documents hierarchically with two models.  First, one model component generates the outline. We then condition upon the outline to generate the entire document. 

Our main contribution will be to explore generating text in a hierarchical manner by separating text generation into two phases. The first phase generates an outline of the text, while the second phase uses the outline to generate the complete article.

\section{Background}

Most sequence-to-sequence models consist of an encoder and decoder model \cite{Sutskever2014}. The encoder model's purpose is to take in an input sequence and construct representations of each token in the sequence. The decoder's hidden state is initialized with encoder's final hidden state. At each step, the decoder predicts the next token in the sequence until it predicts an end of sequence token.

As it is difficult to encode an entire sequence into one vector, generally the decoder is allowed to look back at the representations of the tokens the encoder created through an attention mechanism \cite{Bahdanau2014}. An attention mechanism scores the decoder's hidden state against the encodings of all the tokens in the input sentence, converts these scores to probabilities, and then takes a weighted average of each encoding to produce the context vector. This context vector is then fed in to the decoder to aid it in keeping track of information from the entire sentence.

One weakness of conventional attention is that it can only focus on words in the prompt sequence and not in the target sequence. Self-attention \cite{Vaswani2017} is a modification that attends to both words in the prompt and prior words in the target. This is especially important for models that generate long sequences to be able to keep track of what has been made. A second weakness of conventional attention when used on long sequences of input text is that it only operates at the word level, ignoring any information conveyed through the fact that words are grouped into sentences. An extension of attention to account for both word level and sentence level information is hierarchical attention \cite{Ling2017}, which constructs and scores a vector for each sentence along with each word. The probability of focusing on a word is then the product of its associated word level attention and sentence level attention.

For summarization, approaches fall into two primary categories. Extractive summarization involves choosing and directly copying the main sentences/words from the document. Abstractive summarization focuses on having a model generate the words directly for the summary. As our goal is to simply have an outline of the text, extractive summarization is sufficient. It would be interesting future work to see how different methods of generating an outline affect document generation. Another possible future extension would be to instead use topic modeling to extract a different form of useful context for the target text.

The summarization algorithm that will be used is called SumBasic \cite{Nenkova2005}. This algorithm is based upon choosing sentences with words that are frequent in the document. After choosing a sentence, the algorithm down-weights the words in that sentence to avoid choosing sentences that are too similar.

\section{Methods}

Our goal is to generate text in a hierarchical manner by generating the topic sentences of the article first and then generating the entire paragraph by conditioning on the topic sentence.

\subsection{Model Overview}

\begin{figure}[ht]
\begin{center}
\centerline{\includegraphics[width=\columnwidth]{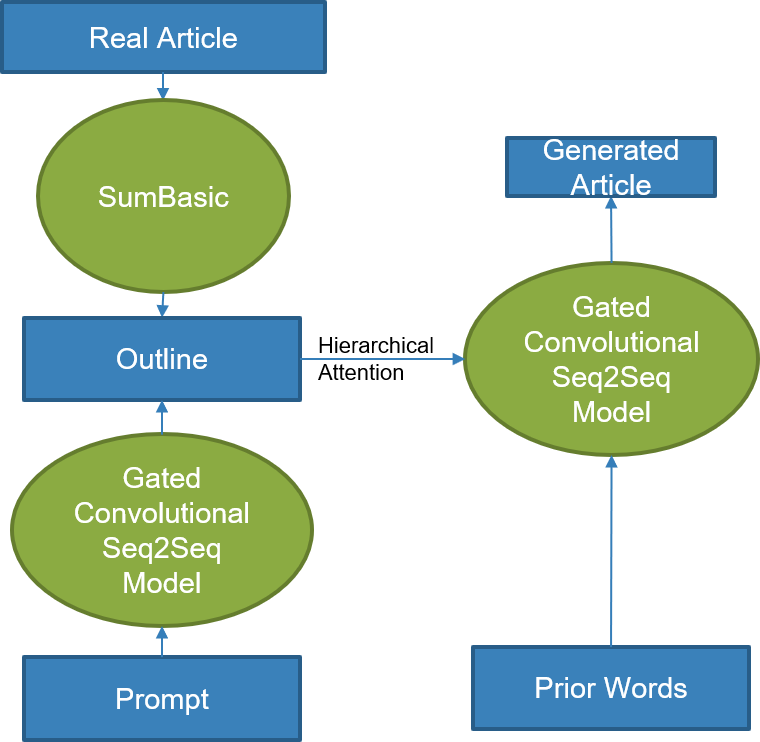}}
\caption{This diagram gives a high level overview of the article generation process. Outlines are prepared with SumBasic. The prompt is converted to an outline which is then converted to an article.}
\label{modelDiagram}
\end{center}
\vskip -0.2in
\end{figure}

The model will be an extension of the model used in Neural Story Generation \cite{Fan2018}. In that work, a sequence-to-sequence model was used to convert a prompt into a complete document. Our main extension is dividing the sequence-to-sequence model into two components to assist in the coherence for the generated text. One component will generate topic sentences, and one will expand topic sentences into complete paragraphs.

\subsection{First Component - Outline Generator}
To generate a document from the prompt, we first generate the outline associated with the document. This component will be trained using almost an architecture almost identical to the model used by \cite{Fan2018}. Their architecture is a convolutional sequence-to-sequence model \cite{Gehring2017} that uses a gated form of self-attention. The one difference is we did not use the cold fusion \cite{Sriram2017} mechanism, mainly to lower training time. Cold fusion is a mechanism that involves first training the main model, freezing its weights, and then training a second randomly initialized version of the main model, concatenating the two models' outputs, and then adding a few layers to determine the final prediction.

\subsection{Second Component - Article Generator}
The second component will use the outline to generate the entire article. It will also be based upon the sequence-to-sequence model used by \cite{Fan2018} and similarly will not include cold fusion. It will be extended in one way. The self-attention heads in the decoder will remain the same, but the attention over the encoded outline will be replaced by hierarchical attention. The sentence vectors will be obtained by summing the encoded word vectors for each word in the corresponding sentence. The sentence vectors will then also have the same type of attention mechanism applied to them. Similar to how the prior decoder attention was gated, the hierarchical attention is also gated and uses the same gating. Both model components will be trained separately by directly optimizing the negative log likelihood.

\subsection{Outlines}
To train the model, we require a dataset of outlines corresponding to complete articles.  As the primary focus of this work is not on new methods of summarization, these outlines will be generated using prior text summarization methods. SumBasic \cite{Nenkova2005} is one frequency-based method for determining the topic sentence. It tries to find sentences whose words are common in the document. One way to avoid overweighting common words like 'the' is to penalize words that are common across many articles. TF-IDF (Term Frequency Inverse Document Frequency) does this by multiplying by the negative log probability of a word appearing in a document. SumBasic is often tweaked to use TF-IDF instead of directly using word frequencies \cite{Allahyari2017}.

As the outline is intended to contain information about each section of the text, instead of directly applying SumBasic to the full document, it will be applied at the meta-paragraph level. Here, meta-paragraph does not refer to the actual paragraphs in the text, because their lengths are very inconsistent. The paragraphs of many of the documents in the training data only consist of one or two sentences. Extracting a topic sentence from each of these paragraphs as an outline would be problematic as we would effectively be letting the outline contain too much of the document. To avoid this issue, the actual paragraphs will be aggregated together using the rule that any paragraph under a threshold number of sentences $k$ will be combined with the next paragraph to form a meta-paragraph. As a side effect, every meta-paragraph except for possibly the last one will end up having at least $k$ sentences.

Lastly, some preprocessing is done before directly applying SumBasic. Stop words are removed using a list of NLTK's \cite{Loper2002} English stop words, numbers are removed, punctuation is removed, and words are stemmed using Porter stemming \cite{Porter1980}.

\subsection{Training the Models}
Strictly speaking the probability of an entire document can only now be obtained by marginalizing over all possible outlines. A lower bound for the probability can be obtained by instead generating the most likely outline for a given prompt and then finding the probability of the document conditioned on that outline. This lower bound however is intractable to compute as it involves generating many outlines.  Due to the long length of the sequences and the relatively slow generation time, even only using 10 outlines for the approximation would take approximately 100 days just to evaluate on the validation/test set on one 1080 Ti. It also turns out to be intractable as the memory needed to do a beam search with such long sequences is too high and it runs out of memory if you increase the beam size beyond 2.

As the two components are trained separately, this similarly leads to the overall training loss not corresponding to the actual article negative log likelihood. Using that loss properly would require being able to efficiently compute the probability of a document, but this is intractable for the reasons given in the prior paragraph. 

A second discrepancy that arises is that since the components are trained  separately, the second component is only trained on good outlines. It's never trained to deal with poorly generated outlines, so if the first component generates a bad outline, the second component is unlikely to be able to recover from the mistake. This is unlike the training for StackGAN \cite{Zhang2016a} where the second part of the model was trained using the output of the first part of the model. Training this model in an end-to-end manner is difficult because the slow outline generation time would increase the training time by a factor of roughly 30 which was unfeasible given our computational resources. Training on a small amount of generated samples to fine-tune the model is feasible and could be done in future work.

\subsection{Code}

A github repository containing our model code can be found here: \href{https://github.com/hmc-cs-mdrissi/fairseq}{\nolinkurl{https://github.com/hmc-cs-mdrissi/fairseq}}.  The dataset of prompts, articles, and outlines can be found at this link: \href{https://www.dropbox.com/s/jupoljc2cx0to7y/datasets.zip?dl=0}{\nolinkurl{https://www.dropbox.com/s/jupoljc2cx0to7y/datasets.zip?dl=0}}.

\section{Evaluation Approach}

\subsection{Datasets}

We used the Wikitext-103 dataset described in \cite{Merity2016}. It consists of a large collection of preprocessed Wikipedia articles. As this dataset does not come with prompts, the prompt used was the first sentence of the article, with the goal being generating the entire article. We preprocessed the dataset to eliminate tables, to canonicalize numbers, and to lowercase each word. 

\subsection{Possible Evaluation Metrics}

The two automated evaluation metrics used previously by \cite{Fan2018} were perplexity and prompt relevance. BLEU and ROUGE were not used as the primary goal is generating new, diverse text rather than matching the target text precisely. Perplexity and prompt relevance are problematic for the hierarchical model as they both involve computing the probability of a article.  Due to the previously discussed computational intractability of computing article probabilities, neither can be used for the hierarchical model. For the non-hierarchical models and the components of the hierarchical model, we can still measure the perplexity to see how well the model captures the text distribution. Perplexity is the exponential of the cross entropy of the model distribution compared to the data distribution and is defined in Equation 1, where $q(x)$ is the probability the model assigns to $x$.

\begin{equation}
perplexity = 2^{-\frac{1}{N}\sum_{i=1}^N \log_2 q(x_i)}
\end{equation}

Recently, \cite{Semeniuta2018} explored various evaluation metrics for language models. One automated metric they explored was the Frechet Infersent Distance (FID). The FID metric is computed by encoding all of the generated and real documents as vectors and then comparing the distributions of these vectors by approximating them as Gaussian. The FID metric is problematic as it requires a model that can encode the meaning of a document well. In their work, they only focused on sentence-level generation and were able to use a model that could create sentence vectors. A second issue with the FID metric is that they did not find it sensitive to word order, which indicates that the metric does not correlate well with human perception of quality.

One, method applicable to any model is human evaluation. We generated about 27 articles for each model using the first sentences of articles in our test set as prompts.  We then had native English speakers evaluate ten articles on both global coherence and overall quality on a 5 point Likert Scale. The questions that were used are shown in Figure \ref{figure:Questions}.

\begin{figure*}
\noindent\fbox{\parbox{\textwidth}{Question 1 (Global Coherence) The article should be well-structured and well-organized. 
The article should not just be a heap of unrelated information, but should build from sentence to sentence and paragraph to paragraph. Abrupt changes in topic without any transitions are problematic.}}

\noindent\fbox{\parbox{\textwidth}{Question 2 (Overall Quality) How realistic is the article as a whole given that the article is meant to be a Wikipedia article.}}
\caption{Questions for Human Evaluation of Generated Articles}
\label{figure:Questions}
\end{figure*}

\subsection{Models}

The models that will be compared are a model from prompt-to-outline, outline-to-article, outline-to-article + hierarchical attention (h.a.), prompt-to-article, hierarchical prompt-to-article + h.a..

\begin{center}
\begin{table}
\caption{Model Perplexity}
\begin{tabular}{|l|l|}
\hline
Model                                & Validation Perplexity \\ \hline
prompt-to-outline         & 45.63         \\ \hline
outline-to-article          & 21.08         \\ \hline
outline-to-article + h.a.   & 20.49         \\ \hline
prompt-to-article           & 30.96         \\ \hline
\end{tabular}
\end{table}
\end{center}

\section{Evaluation Results}
\subsection{Perplexity Evaluation}

The model perplexity experiments revealed multiple interesting findings. The biggest one is that the perplexity of prompt-to-article is lower than the perplexity of prompt-to-outline, indicating that it is easier on average word-wise to generate the full article than just the outline. One possible explanation for this is that in generating the outline there are much more abrupt shifts in topic when compared to generating the article, and as each word is conditional on the prior words, the topic flowing more smoothly may make it easier to guess the next word. There is a significant improvement in perplexity (about 10 points) when conditioning on outlines compared to conditioning on prompts.

\begin{table*}
\label{table:human_results}
\begin{center}
\begin{tabular}{lcccc}
\hline
Model & $\mu$-Global Coherence & $\sigma$-Global Coherence &  $\mu$-Quality & $\sigma$-Quality \\ \hline
prompt-to-article                       &  3.36   & 1.00   &  2.91    &  1.07  \\
hierarchical prompt-to-article + h.a.   &  2.54   & 0.96   &  2.26    &  0.89  \\  
outline-to-article + h.a.               &  3.14   & 1.31   &  2.90    &  1.22  \\ 
\hline
\end{tabular}
\end{center}
\caption{The $\mu$ is for mean and $\sigma$ is for standard deviation. }
\end{table*}

The other interesting result is that hierarchical attention led to a small improvement in perplexity. In the prior work by \cite{Ling2017}, hierarchical attention improved the model by providing a  computationally efficient attention mechanism over long sequences and was not found to be helpful otherwise. A similar magnitude improvement using hierarchical attention occurred when comparing the two models on the stories dataset used by \cite{Fan2018}. Two plausible explanations are either that hierarchical attention works better for some tasks/datasets, or that the discrepancy occurred because we used gated attention for both the word-level attention model and the hierarchical attention model. The gating may have helped the sentence attention learn to focus on different and useful things than the word attention. Lastly, the hierarchical model is missing from the table as computing perplexity for it was intractable.

\subsection{Human Evaluation}

The results can be found in Table \ref{table:human_results} on the next page. The number of articles evaluated total is 70. For the three models, only one pair is significantly different in global coherence. That pair is hierarchical model vs prompt-to-article where the latter is better ($p = 0.005$). The hierarchical model is also significantly different in quality from both of the other two models (for both $p = 0.049$). The worst model is still within a standard deviation evaluation wise of the best model in both metrics, indicating how noisy the quality of the articles and the reviews of them are. Thus, for human perception all the hierarchical model performed slightly worse, while the outline vs prompt models performed equally well.

The biggest unexpected result is that the prompt-to-article model performed better in human evaluation on global coherence than the outline-to-article model. Considering the outline describes the article more thoroughly, it should have helped the model stay focused. It was also unexpected that the overall quality of the outline-to-article would be the same as the prompt-to-article model. Both of these results, are unexpected because there was a large improvement in perplexity when conditioning on an outline. This reveals that for this task, perplexity did not correlate well with human perception of quality. Lastly, overall quality and global coherence correlate strongly ($r=0.74$, $p < 0.001$).

\subsection{Best Cases}
Due to space constraints, articles and outlines were both truncated. unk was used for unknown tokens, num was used for numbers, and newline was used to keep track of paragraph breaks.  The article shown in Figure \ref{figure:BestArticle} came from the outline-to-article model and tied for highest human evaluation scores.

\begin{figure*}
\vspace*{.15in}
\noindent\fbox{\parbox{\textwidth}{Outline 1: he was previously the head coach at butler university in indianapolis . newline on april num , num , he became the head coach after todd unk left to coach the iowa hawkeyes . newline in july num , he signed a six - year , num million dollar contract to become the head coach of the boston celtics in the nba . newline  bradley kent stevens grew up in unk , indiana , where he developed his love for basketball . newline he wore no. num in high school in honor of indiana pacers guard reggie miller . newline he earned three letters in basketball , three in track , and one in baseball during his days at unk . newline the more success you had at lilly , the harder it would be to leave . "}}

\noindent\fbox{\parbox{\textwidth}{Generated Article 1: walter george stevens ( may num , num october num , num ) was an american major league baseball ( mlb ) outfielder who played for the boston red sox ( num \ num , num  num ) and the yankees ( num  num ) . stevens was an outfielder , placekicker and punter , and had played with the boston indians and boston red sox from num to num . he was previously the head coach at butler university in indianapolis . he played num seasons in major league baseball ( mlb ) for the boston red sox , and in num played on the chicago red sox from num to num . newline stevens played num major league baseball ( mlb ) all - star team in num , and led the league to five consecutive national champions with a num  num conference record , the longest streak in career history . he was also the only player to reach the major division i 's num playoffs . the num and num seasons have taken place in the central division ( num  num ) . newline at the time of the num mlb draft , he was known for the right to handle a hit . he was one to only two players . in num , the red sox traded him in minor league organization with the red sox . the red sox promoted him with the numth overall pick in the num mlb draft , while the red sox traded him to the cleveland indians .}}
\caption{Well-generated Article}
\label{figure:BestArticle}
\end{figure*}

In both cases we see that the text conditioned upon was very informative. Doing a better job of preprocessing would likely help in some cases, as some of the other prompts end up being truncated too heavily (mainly due to difficulty determining where a sentence ends). While the main topic is preserved, often the facts will end up being changed. For the second article, originally the player was a basketball player, but the model morphed him into a baseball player.

\subsection{Failure Cases}

Here, we will examine one of the worst generated articles. Worst is defined as being rated as a 1 in both global coherence and overall quality. The article, shown in Figure \ref{figure:PoorArticle} was generated from the hierarchical model. 

\begin{figure*}
\noindent\fbox{\parbox{\textwidth}{Prompt 1: the portage to san cristobal of a.h. is a num literary and philosophical novella by george steiner , in which jewish nazi hunters find a fictional adolf hitler ( a.h. ) alive in the amazon jungle thirty years after the end of world war ii .}}

\noindent\fbox{\parbox{\textwidth}{Generated Outline 1: he began work as a unk at the age of num . newline during world war ii , unk joined the unk unk ( royal society ) in the united states where he made his world debut in the world . newline unk 's first name was unk unk . newline he was promoted to unk 's unk unk - unk in num . newline unk became the new man 's second - oldest student , unk unk - unk , as a student from unk - unk . newline unk was one of the founding members of the unk unk - unk , the first to be called unk .}}

\noindent\fbox{\parbox{\textwidth}{Generated Article 1: unk unk ( unk , " young woman 's sister , the unk , " to the unk " ) is unk ( unk , " young woman 's sister , the unk ) , " girl 's son , and the unk " . the name unk is from the greek unk or " woman 's mother unk " , referring to her marriage to unk . after the marriage , her son , ila , is the only person to be killed in the world . newline during world war ii , il began work as a unk at the age of num . she was the first female woman to be a unk ( unk ) . after being married , ila became the new man 's second - oldest student ; however , her marriage to unk was interrupted in num . newline during world war ii , il joined the unk unk ( royal society ) in the united states where he made his world debut in the world . the unk - unk unk ( unk ) , the first of the first unk ( unk ) , was the first unk to be a unk , the second unk in the world , and a member of the unk unk , one of the first women 's women 's women 's women and is the first female woman to be a unk . in her unk unk , ila was first unk unk unk ; the other two are the unk ( unk ) , the unk ( unk ) , the ( unk ) , the unk ( unk ) , the unk ( unk ) , the unk ( unk ) , the unk ( unk ) , the unk ( unk ) and the ( unk ) . ila 's first name was unk unk . she was also used to refer to her .}}
\caption{A Poorly Generated Article}
\label{figure:PoorArticle}
\end{figure*}

This example looks to be a case where the first component of the hierarchical model did poorly leading to the model struggling to generate a article from that outline. It is not the only example where too many unknown word tokens end up leading to poor articles.  While repetition occurs more broadly, unknown tokens have the worst tendency to lead the article to become too repetitive. This issue may be helped by training the second component of the hierarchical model on not just good outlines, but also sampled outlines so it can learn to not be too reliant on the outline.

More broadly, the main type of error found in generated articles is too much redundancy. Often individual sentences would not be too bad, but a very similar sentence would appear a bit later in the paragraph. Article quality also generally becomes worse near the end of the article.

\section{Conclusion}

We have focused on generating text in a hierarchical manner for generating Wikipedia articles. The primary automated metric, perplexity, for measuring overall article generation quality was computationally intractable for the hierarchical model. We found that conditioning on an outline heavily improved perplexity, but did not improve human perceived quality. This indicates that, perplexity should be used more cautiously for text generation if the motivation is generating text people will find believable. This issue should be analyzed more thoroughly in future work as perplexity is the most commonly used evaluation metric for language models currently. More generally, it would be useful to develop automated metrics that more closely correlate with human evaluation and have a better understanding of what qualitative aspects good articles share.

Other interesting findings are that that one major failure mode for the hierarchical model is difficulty dealing with poor outlines and that hierarchical attention led to a small improvement in perplexity for two different datasets unlike when it was used previously by \cite{Ling2017}.

For future work, the main thing is to deal with the issue of the hierarchical model not being able to generate reasonable articles with poor outlines. Training using only generated articles would be ideal, but is computationally expensive due to generation time being hundreds of times slower than training time. To make training feasible with generated samples, it will likely require only using a small number of generated articles to help fine tune the model. This may also be feasible when the training dataset size is small.

Another possible
extension is to make better usage of the outlines. Specifically, for ideal outlines we know that the sentences present in the outline should also be present in the target text. Allowing the model to copy an entire sentence would be beneficial and make it more likely to fully utilize the outline. This does come with the downside that a poor sentence in a generated outline being copied would be problematic. That could be dealt with by training on a mixture of generated and real outlines.

\section{Acknowledgements}
I would like to thank the UCCS NLP lab for many helpful conversations. This material is based upon work supported by the National Science Foundation under Grant No. 1659788.  Any opinions, findings, and conclusions or recommendations expressed in this material are those of the author(s) and do not necessarily reflect the views of the National Science Foundation.

\balance
\bibliography{all_papers_6_1_18}
\bibliographystyle{acl_natbib}

\end{document}